# `seqme`: a Python library for evaluating biological sequence design

Rasmus Møller-Larsen[1,2], Adam Izdebski[1,2], Jan Olszewski[3], Pankhil Gawade[1], Michal Kmicikiewicz[1,2], Wojciech Zarzecki[3,4] and Ewa Szczurek[1,3,*]

[1]Institute of AI for Health, Helmholtz Munich, Ingolstädter Landstraße 1, 85764, Neuherberg, Germany, [2]School of Computation, Information and Technology, Technical University of Munich, Boltzmannstraße 3, 85748, Garching bei München, Germany, [3]Faculty of Mathematics, Informatics and Mechanics, University of Warsaw, Stefana Banacha 2, 02-097, Warszawa, Poland and [4]Faculty of Electronics and Information Technology, Warsaw University of Technology, Nowowiejska 15/19, 02-097, Warszawa, Poland

*Corresponding author. ewa.szczurek@helmholtz-munich.de

**Abstract**

**Summary:** Recent advances in computational methods for designing biological sequences have sparked the development of metrics to evaluate these methods performance in terms of the fidelity of the designed sequences to a target distribution and their attainment of desired properties. However, a single software library implementing these metrics was lacking. In this work we introduce `seqme`, a modular and highly extendable open-source Python library, containing model-agnostic metrics for evaluating computational methods for biological sequence design. `seqme` considers three groups of metrics: sequence-based, embedding-based, and property-based, and is applicable to a wide range of biological sequences: small molecules, DNA, ncRNA, mRNA, peptides and proteins. The library offers a number of embedding and property models for biological sequences, as well as diagnostics and visualization functions to inspect the results. `seqme` can be used to evaluate both one-shot and iterative computational design methods.

**Availability and implementation:** `seqme` is released at https://github.com/szczurek-lab/seqme under the BSD 3-Clause license.

**Supplementary information:** Supplementary Material is available at *Bioinformatics* online.

## Introduction

Biological sequences are composed of ordered chains – of nucleotides in DNA and RNA, amino acids in peptides and proteins, and atoms in small molecules. These sequences carry the information on structure, interactions, and function of the encoded molecules, making sequence modeling central to understanding and engineering biology (Kozak, 1986; Anfinsen, 1973). Recent years have witnessed a surge of generative AI models and algorithms for biological sequence design, including the development of novel compounds (Tang et al., 2024; Izdebski et al., 2025), peptides (Szymczak et al., 2025; Soares et al., 2025), and proteins (Kortemme, 2024; Kmicikiewicz et al., 2025). Evaluating biological designs requires multiple complementary criteria, such as fidelity to a reference dataset's distribution, novelty, diversity and optimization of desired properties. Overlooking these aspects when training computational methods, in particular generative AI models, can result in failures to generate sequences that are both high quality and functionally useful (Theis et al., 2016). Therefore, a plethora of evaluation metrics have been proposed to measure performance and identify failure modes of generative models (Manduchi et al., 2025). However, to date, no software library has been developed that unifies these efforts and is dedicated specifically to evaluating biological sequence design.

To address this gap, we introduce `seqme` - the first library for end-to-end evaluation of generative AI and computational algorithms for *de novo* biological sequence design. The library provides access to a collection of metrics, embedding models and property predictors, while remaining highly extensible. It supports major types of biological sequences, including small molecules, DNA, ncRNA, mRNA, peptides and proteins. In addition, `seqme` enables assessment of not only single-shot design but also iterative sequence discovery pipelines. By offering a unified and reproducible framework, the library establishes a practical foundation for fair comparison and accelerated progress in biological sequence design.

## Metrics

`seqme` offers metrics for measuring *distribution*- and *property*-based quality of designed sequences, i.e., the degree to which the sequences resemble a biological reference distribution and to which they satisfy pre-specified target properties or functions. Depending on whether the metrics operate directly on raw sequences, on their learned embeddings, or on derived properties, we further categorize them into three classes: sequence-, embedding- and property-based metrics, respectively. The currently implemented

**Table 1.** Metrics currently implemented in `seqme`. Arrows indicate whether a metric ought to be maximized (↑) or minimized (↓). (D) and (P) specify whether a metric measures *distribution-* or *property*-based sequence quality. For an up-to-date list, check https://seqme.readthedocs.io/en/stable/api/metrics_index.html.

| | Metric | Description |
|---|---|---|
| **Sequence** | Novelty (↑) (D) | Fraction of sequences not in the reference set. |
| | Uniqueness (↑) (D) | Fraction of unique sequences. |
| | Diversity (↑) (D) | Normalized Levenshtein distances between sequences. |
| | N-gram Jaccard Similarity (↑/↓) (D) | Jaccard similarity between sequences and reference set's n-gram. |
| **Embedding** | Fréchet Biological Distance (FBD) (Heusel et al., 2018) (↓) (D) | 2-Wasserstein distance between two multivariate Gaussian distributions fitted to the embeddings of the sequences and reference set. |
| | Maximum Mean Discrepancy (MMD) (↓) (D) | MMD between the sequences and references embeddings using the Gaussian RBF kernel (Jayasumana et al., 2024) or the rational quadratic kernel (Bińkowski et al., 2021). |
| | Precision (Kynkäänniemi et al., 2019) (↑) (D) | Fraction of sequences that fall inside the support of the reference set. |
| | Recall (Kynkäänniemi et al., 2019) (↑) (D) | Fraction of references that fall inside the support of the sequences. |
| | Authenticity (Alaa et al., 2022) (↑) (D) | Fraction of sequences whose nearest training neighbor is closer to some other training sample than to the sequence. |
| | Vendi score (Friedman and Dieng, 2023) (↑) (D) | A reference-free method to estimate diversity. |
| **Property** | Identity (ID) (↑/↓) (P) | Mean and standard deviation of a single property across the sequences. |
| | Threshold (↑/↓) (P) | Fraction of sequences where a user-defined property is above (or below) a threshold. |
| | Hit-rate (↑) (P) | Fraction of sequences satisfying a user-defined set of conditions. |
| | Hypervolume (↑) (P) | Hypervolume of two or more properties derived from the sequences. Computed either as the hypervolume indicator (Zitzler and Thiele, 1999) or the convex-hull. |
| | Conformity score (Frey et al., 2024) (↑) (D) | Distributional similarity of the sequences properties and the references properties. |
| | KL-divergence (↓) (D) | Kullback–Leibler divergence between sequences and references for a single property. |

metrics are summarized in Table 1. Formal definitions are found in Supplementary Text.

### Sequence-based metrics

`seqme` includes the metrics *novelty*, *diversity*, and *uniqueness*, which are commonly used to detect failure modes. There is a tradeoff between maximizing these metrics and optimizing user-defined properties. A typical failure mode is overfitting to a small set of biological sequences that satisfy the target properties at the expense of novelty, diversity, and uniqueness. Conversely, a random baseline can trivially maximize novelty and diversity without actually optimizing the desired properties, since the sequence space in biological domains is enormous, e.g., peptide space comprises approximately $20^{50} \approx 1.13\mathrm{e}65$ possible sequences. Thus, these metrics are necessary for a comprehensive evaluation of biological sequence design methods.

### Embedding-based metrics

`seqme` implements the metric *Fréchet Biological Distance* (FBD) (Heusel et al., 2018), which is commonly used to evaluate the distributional similarity between biological sequences designed by a model and those in a reference dataset (Stark et al., 2024; Preuer et al., 2018). However, FBD is sensitive to both the number of sequences and their distribution in the embedding space. Therefore, `seqme` also includes *Maximum Mean Discrepancy* (MMD), which provides more stable estimates (Jayasumana et al., 2024). In addition, the library implements reference-free metrics such as *Vendi-* and *RKE-score* (Friedman and Dieng, 2023; Ospanov et al., 2024) to evaluate diversity in the embedding space. None of the aforementioned metrics can identify memorization, i.e., near-sequence copying in the embedding space. To address this, `seqme` includes the *Authenticity* metric (Alaa et al., 2022). Finally, the library also implements *Improved Precision* and *Improved Recall* (Kynkäänniemi et al., 2019) to evaluate fidelity and diversity in the embedding space.

### Embedding models

Recall that embedding-based metrics rely on embedding models that map sequences to fixed-length vector representations. For protein embeddings, `seqme` includes ESM-2 (Lin et al., 2022), for peptides it contains ESM-2 finetuned on peptide sequences and Hyformer trained on peptides (Izdebski et al., 2025), for ncRNA and mRNA, it uses RNA-FM (Chen et al., 2022), for DNA it includes GENA-LM (Fishman et al., 2025) and for small molecules Hyformer trained on small molecules (Supplementary Table 1). The library also supports the use of alternative embedding models.

Selecting an appropriate embedding model is a non-trivial task as the embeddings must capture the biological domain of interest. To assist with this, `seqme` provides functionality for visualizing the embedding space using PCA, t-SNE, and UMAP 2D projections. Furthermore, `seqme` offers diagnostic tools, such as the k-nearest neighbor feature-alignment score and Spearman alignment score (Rissom et al., 2025), to evaluate how well embedding models align with discrete or continuous sequence properties of interest (See Supplementary Text for more details).

### Property-based metrics

`seqme` implements, among others, the metrics *Conformity Score* (Frey et al., 2024) and *Hit Rate*. The latter is commonly used in drug discovery pipelines to quantify the fraction of sequences that satisfy desired properties. In addition, `seqme` includes the multi-objective optimization metric *Hypervolume Indicator* (Zitzler and Thiele, 1999), which computes the hypervolume of two or more properties across a set of sequences.

### Property models

Property-based metrics assume access to property models. `seqme` provides functionality to compute several physico-chemical properties of proteins and peptides, as well as models predicting whether a peptide exhibits antimicrobial activity (see Supplementary Table 1). Furthermore, `seqme` integrates ESMFold (Lin et al., 2022), a model trained on protein sequences to predict their three-dimensional structure. Finally, the library contains models to predict DNA promoter regions and splice sites.

## Additional functionalities of `seqme`

### Fold functionality

Several metrics, such as *Improved Precision* and *Improved Recall* (Kynkäänniemi et al., 2019), can become biased when there is a discrepancy in the number of designed sequences and reference sequences. Furthermore, for any evaluated metric, it is often desirable to estimate its standard deviation across multiple subsets of the designed sequences. To address these matters, we introduce *Fold*, which splits the sequences into $K$ groups of equal size and computes the metric $K$ times. This approach mitigates metric-specific sample-size biases, reduces the number of discarded sequences, and enables the estimation of standard deviation or standard error for metrics.

### Support of iterative designs

As its core functionality, `seqme` enables evaluation of a given tuple of sequences using a set of metrics of interest. This is convenient for one-shot design, where the given tuple of sequences is evaluated only once. On top of that, `seqme` also supports iterative evaluation workflows. This functionality enables the same metrics to be applied iteratively, for example to evaluate design quality across training epochs or iterations of generative models, or throughout the iterative sequence refinement process in genetic algorithms and Bayesian optimization.

### Visualizations

`seqme` provides several convenient functions for visualizing metric results. It can display a customizable table of metrics for all evaluated sequence groups and plot a parallel-coordinates chart, where each axis corresponds to one metric. For inspecting a single metric, `seqme` offers a barplot with optional error or deviation bars. For iterative sequence design, it provides functionality to plot metric trajectories across design iterations for multiple sequence designs. These visualizations make sequence-design performance easy to interpret and help reveal potential failure modes.

## Implementation

`seqme` is implemented as a highly extendable Python library (v3.10 or greater) while maintaining a simple interface (see *Example usage* box). `seqme` offers capabilities for performance optimization by caching sequence representations, i.e., sequence embeddings and properties. Caching provides a substantial speed-up when using several metrics with the same sequence embeddings and the embeddings models are large, e.g., most language models. With caching enabled, the time-complexity of computing $m$ metrics with the same embedding model for $n$ sequences is reduced to $\mathcal{O}(n + k)$ instead of $\mathcal{O}(nk)$, where $k = \sum_{i=1}^{m} m_i$. `seqme` contains extensive documentation and tutorials on how to use the library and how to add new metrics and models. Notably, we show how to integrate models with package dependency conflicts using `seqme`'s third-party model interface. More details can be found in the Supplementary Text. We expect that both the number of metrics and models supported by the library will grow as the community adopts the library.

**Example usage**

```python
import seqme as sm

sequences = {
    "UniProt": ["GFGD", "DPWDWV", "IEFFT"],
    "DBAASP": ["PGLGFY", "AAVLNA", "LAHRYH"],
}

cache = sm.Cache(
    models={
        "esm2": sm.models.ESM2(
            model_name="facebook/esm2_t6_8M_UR50D",
            batch_size=256, device="cpu",
        ),
    }
)

metrics = [
    sm.metrics.Diversity(),
    sm.metrics.FBD(
        reference=sequences["UniProt"],
        embedder=cache.model("esm2")
    ),
]

df = sm.evaluate(sequences, metrics)
sm.show(df)
```

## Discussion

`seqme` aims to ease the task of evaluating biological sequence design methods by providing metrics, as well as embedding and property models. While the library is already applicable to multiple biological sequence types, it is easily extendable to more types by adding type-specific embedding and property models.

We encourage users to use multiple metrics provided by `seqme`, as each metric offers a different perspective. This is of particular importance since each evaluation metric is imperfect and can fail to detect a failure in a given generative AI model or algorithm used for the design (Räisä et al., 2025). As different metrics have different limitations, only the use of several carefully chosen metrics will yield a robust evaluation.

We envision `seqme` will accelerate *de novo* drug discovery by providing the foundation for robust model benchmarking, and offering more comprehensive tools for defining stopping criteria in machine-learning training loops. Ultimately, we expect that libraries like `seqme` will enable practical applications of generative AI and other sequence design methods, and facilitate the translation of computational advances into biology and medicine.

## Competing interests

Projects at Szczurek lab at the University of Warsaw are co-funded by Merck Healthcare.

## Author contributions statement

R.M.-L., A.I., and E.S. conceptualized the project and wrote the manuscript. R.M.-L., A.I., J.O., P.G., M.K., and W.Z. developed the software. All authors contributed to the Supplementary Material.

## Acknowledgments

This work was supported by the European Research Council (ERC) under the European Funding Union's Horizon 2020 research and innovation programme, grant no. 810115. We thank Bruno Puczko-Szymanski, Serra Korkmaz and Jurand Pradzynski for contributing to the library during the Szczurek lab hackathon.

## Code availability

seqme is installable using `pip install seqme` through PyPI. The GitHub repository is available at https://github.com/szczurek-lab/seqme. Documentation and tutorials are available at https://seqme.readthedocs.io. Third-party models are available at https://github.com/szczurek-lab/seqme-thirdparty. Library dependencies and their licenses can be found in the Supplementary Text.

# `seqme`: a Python library for evaluating biological sequence design Supplementary Material

Rasmus Møller-Larsen, Adam Izdebski, Jan Olszewski,
Pankhil Gawade, Michal Kmicikiewicz, Wojciech Zarzecki,
Ewa Szczurek

# 1 Supplementary Text

## 1.1 Definitions

### 1.1.1 Data

In the definitions of metrics measuring performance of generative models, we will consistently refer to the following three objects:

1. $\mathcal{G}$ – set of generated outputs from the model.
2. $\mathcal{R}$ – set of reference data, which usually comes from the model's training or evaluation dataset.
3. $\varphi$ – representation function, which usually decodes data to sequence strings or embedding vectors.

We assume that both the reference and generated sets share a common domain $\mathcal{D} \supset \mathcal{R}, \mathcal{G}$.

It is worth mentioning that we consider the elements $x, y \in \mathcal{G}$ different as long as they were generated by two distinct calls to the generative model, even if their string representations are identical

$$\varphi_{\text{string}}(x) = \varphi_{\text{string}}(y) \not\Longrightarrow x = y.$$

The same holds for elements of the reference set $\mathcal{R}$.

### 1.1.2 Representation functions

We distinguish between three categories of representation functions $\varphi$:

1. **String representation** which for each element assigns its decoded sequence. The sequence is a list of $n$ letters coming from a finite alphabet $\mathcal{A}$. In the case of peptide sequences, $|\mathcal{A}| = 20$ is the alphabet of canonical amino acids. Formally

$$\varphi_{\text{string}} : \mathcal{D} \to \bigcup_{n=1}^{\infty} \mathcal{A}^n, \text{ where } \mathcal{A}^n = \{(a_1, a_2, \ldots, a_n) : a_i \in \mathcal{A}\}.$$

2. **Fixed-size embeddings** which employ some embedding model $\mathrm{E} : \varphi_{\text{string}}(\mathcal{D}) \to \mathbb{R}^d$, usually a pretrained deep learning model, to map the variable-length sequences to a fixed $d$-dimensional Euclidean space. Formally

$$\varphi_{\mathrm{E}} : \mathcal{D} \to \mathbb{R}^d, \text{ where } \varphi_{\mathrm{E}}(x) = \mathrm{E}(\varphi_{\text{string}}(x)).$$

3. **Sequence property** in which for each function $\mathrm{P} : \varphi_{\text{string}}(\mathcal{D}) \to Y$ assigning some property to a sequence, we define

$$\varphi_{\mathrm{P}} : \mathcal{D} \to Y, \text{ where } \varphi_{\mathrm{P}}(x) = \mathrm{P}(\varphi_{\text{string}}(x)).$$

Assumptions on the co-domain $Y$ will be introduced separately for each metric definition.

### 1.1.3 Metric

A metric is a function $m$ of generated data $\mathcal{G}$, which scores its quality

$$m : \mathcal{P}(\mathcal{D}) \to \mathbb{R}, \text{ where } \mathcal{P}(\mathcal{D}) = \{\mathcal{G} : \mathcal{G} \subset \mathcal{D}\}.$$

Optimizing the metric is desired, and the optimal set $\mathcal{G}^*$ is either minimizer $m(\mathcal{G}^*) = \inf_{\mathcal{G}}\{m(\mathcal{G})\}$ or maximizer $m(\mathcal{G}^*) = \sup_{\mathcal{G}}\{m(\mathcal{G})\}$ of the metric.

## 1.2 Metrics

In this section, we define metrics to evaluate generated data $\mathcal{G}$ based on reference data set $\mathcal{R}$ and representation function $\varphi$. All metrics are implemented in the `seqme` library. The metrics naturally split into three categories depending on the employed function $\varphi$ to represent the data.

### 1.2.1 Metrics from string representation

In this group, metrics use the representation function $\varphi_{\text{string}}$ which assigns a decoded sequence to each element $x \in \mathcal{D}$. We denote these metrics as sequence-based metrics.

**Novelty** is the fraction of generated data not included in the reference set

$$\text{Novelty}_{\mathcal{R}}(\mathcal{G}) = \frac{|\mathcal{G} \setminus \mathcal{R}|}{|\mathcal{G}|}. \quad (1)$$

**Uniqueness** is the fraction of unique sequences in the generated data

$$\text{Uniqueness}(\mathcal{G}) = \frac{|\varphi_{\text{string}}(\mathcal{G})|}{|\mathcal{G}|}. \quad (2)$$

**Diversity** measures the average pairwise dissimilarity between sequences in $\mathcal{G}$

$$\text{Diversity}(\mathcal{G}) = \frac{1}{|\mathcal{G}|} \sum_{x \in \mathcal{G}} \frac{1}{|\mathcal{G}|} \sum_{y \in \mathcal{G}} \overline{\text{lev}}(\varphi_{\text{string}}(x), \varphi_{\text{string}}(y)), \quad (3)$$

where $\overline{\text{lev}}(a, b)$ is the Levenshtein distance between strings normalized by the maximum string length

$$\overline{\text{lev}}(a, b) = \frac{\text{lev}(a, b)}{\max(\text{len}(a), \text{len}(b))}.$$

Because the cost of computing the Diversity metric grows quadratically in the size of the generated set $O(|\mathcal{G}|^2)$, we approximate the inner sum by sampling a random subset of $k$ elements from $\mathcal{G}$ which does not contain $x$

$$\mathcal{G}_k(x) \sim \text{Unif}[\{\mathcal{G}' \subset \mathcal{G} \setminus \{x\} : |\mathcal{G}'| = k\}].$$

Therefore we get a $k$-approximation of Diversity

$$\text{Diversity}_k(\mathcal{G}) = \frac{1}{|\mathcal{G}|} \sum_{x \in \mathcal{G}} \frac{1}{k} \sum_{y \in \mathcal{G}_k(x)} \overline{\text{lev}}(\varphi_{\text{string}}(x), \varphi_{\text{string}}(y)). \quad (4)$$

The cost of computing $\text{Diversity}_k$ is $O(nk)$. Note that $\text{Diversity}(\mathcal{G}) = \text{Diversity}_{|\mathcal{G}|-1}(\mathcal{G})$.

**N-gram Jaccard similarity** computes the average intersection over union for N-grams of generated data $\mathcal{G}$ and aggregated N-grams of reference data $\mathcal{R}$

$$\text{Jaccard}_{N,\mathcal{R}}(\mathcal{G}) = \frac{1}{|\mathcal{G}|} \sum_{x \in \mathcal{G}} \frac{\text{Ngram}(\varphi_{\text{string}}(x)) \cap \overline{\text{Ngram}}(\varphi_{\text{string}}(\mathcal{R}))}{\text{Ngram}(\varphi_{\text{string}}(x)) \cup \overline{\text{Ngram}}(\varphi_{\text{string}}(\mathcal{R}))}, \quad (5)$$

where the Ngram of a string is a set of its $N$-element substrings. Formally, for $s = \varphi_{\text{string}}(x)$ we define

$$\text{Ngram}(s) = \{s_{i:i+N} : 0 \leq i \leq \text{len}(s) - N\},$$

and for a set of strings $R = \varphi_{\text{string}}(\mathcal{R})$ the aggregated $\overline{\text{Ngram}}$ is defined as

$$\overline{\text{Ngram}}(R) = \bigcup_{s \in R} \text{Ngram}(s).$$

### 1.2.2 Metrics from fixed-size embeddings

In this group, metrics use representation functions $\varphi_E$ that assign a fixed-size $d$-dimensional embedding vector to each element $x \in \mathcal{D}$. We denote these metrics as embedding-based metrics.

**Fréchet Biological Distance (FBD)** is defined identically to the *Fréchet Inception Distance* (FID) (Heusel et al., 2018), but replaces the Inception network with any biologically relevant embedding model E.

Assume that the embeddings of the generated data and the embeddings of the reference data follow a multivariate Gaussian distribution

$$\varphi_E(\mathcal{G}) \sim \mathcal{N}(\mu_\mathcal{G}, \Sigma_\mathcal{G}),$$
$$\varphi_E(\mathcal{R}) \sim \mathcal{N}(\mu_\mathcal{R}, \Sigma_\mathcal{R}).$$

The FBD between the generated data embeddings and the reference data embeddings is defined as

$$\text{FBD}_\mathcal{R}(\mathcal{G}) = \|\mu_\mathcal{G} - \mu_\mathcal{R}\|_2^2 + \text{Tr}\left(\Sigma_\mathcal{G} + \Sigma_\mathcal{R} - 2(\Sigma_\mathcal{G}\Sigma_\mathcal{R})^{1/2}\right), \tag{6}$$

where $\|\cdot\|_2$ denotes the Euclidean norm, $\text{Tr}(\cdot)$ denotes the trace operator, and $(\Sigma_\mathcal{G}\Sigma_\mathcal{R})^{1/2}$ denotes the matrix square root of the product $\Sigma_\mathcal{G}\Sigma_\mathcal{R}$. In practice, we estimate $\mu_\mathcal{G}, \mu_\mathcal{R}, \Sigma_\mathcal{G}, \Sigma_\mathcal{R}$ from the data.

**Maximum Mean Discrepancy (MMD)** together with a positive-definite kernel $\kappa$, is used to compare the distributions of reference data embeddings and generated data embeddings. Assume that $\varphi_E(\mathcal{G})$ and $\varphi_E(\mathcal{R})$ are sampled from corresponding probability distributions $\lambda_\mathcal{G}, \lambda_\mathcal{R}$

$$\varphi_E(\mathcal{G}) \sim \lambda_\mathcal{G},$$
$$\varphi_E(\mathcal{R}) \sim \lambda_\mathcal{R}.$$

Then, the MMD between the generated data embeddings and the reference data embeddings is defined as

$$\text{MMD}_{\kappa,\mathcal{R}}(\mathcal{G}) = \mathbb{E}_{X,X'\sim\lambda_\mathcal{G}}[\kappa(X,X')] + \mathbb{E}_{Y,Y'\sim\lambda_\mathcal{R}}[\kappa(Y,Y')] - 2\mathbb{E}_{X\sim\lambda_\mathcal{G},Y\sim\lambda_\mathcal{R}}[\kappa(X,Y)], \tag{7}$$

where $\kappa : \mathbb{R}^d \times \mathbb{R}^d \to \mathbb{R}$ is a positive-definite kernel function. The usual choice of $\kappa$ is the Gaussian RBF kernel $\kappa(x,y) = \exp(-\|x-y\|_2^2/(2\sigma^2))$ (Jayasumana et al., 2024) or the rational quadratic kernel $\kappa(x,y) = \left(1 + \frac{\|x-y\|_2^2}{2\alpha}\right)^{-\alpha}$, where $\alpha > 0$ (Bińkowski et al., 2021).

In practice, we have access to i.i.d. samples from distributions $\lambda_\mathcal{G}$ and $\lambda_\mathcal{R}$

$$\{X_1, X_2, \ldots, X_n\} \overset{\text{i.i.d.}}{\sim} \lambda_\mathcal{G},$$
$$\{Y_1, Y_2, \ldots, Y_m\} \overset{\text{i.i.d.}}{\sim} \lambda_\mathcal{R},$$

and thus compute an unbiased estimator $\overline{\text{MMD}}_{\kappa,\mathcal{R}}$ of $\text{MMD}_{\kappa,\mathcal{R}}$

$$\overline{\text{MMD}}_{\kappa,\mathcal{R}}(\mathcal{G}) = \frac{1}{n(n-1)}\sum_{i=1}^n\sum_{j\neq i}^n \kappa(X_i, X_j) + \frac{1}{m(m-1)}\sum_{i=1}^m\sum_{j\neq i}^m \kappa(Y_i, Y_j) - \frac{2}{nm}\sum_{i=1}^n\sum_{j=1}^m \kappa(X_i, Y_j). \tag{8}$$

**Improved Precision and Recall** uses intra-set $k$-th nearest neighbor distance to calibrate the binary classification of a point as a set member. We classify $x$ as being included in set $\mathcal{R}$ if there exists $y \in \mathcal{R}$ such that $x$ is closer to $y$ than the $k$-th nearest neighbor of $y$ from set $\mathcal{R}$. We follow Kynkäänniemi et al. (2019) and define the Improved Precision and Recall to measure similarity between sets $\varphi_E(\mathcal{G})$ and $\varphi_E(\mathcal{R})$ as

$$\text{Precision}_{k,\mathcal{R}}(\mathcal{G}) = \frac{|\{x \in \mathcal{G} : \exists_{y\in\mathcal{R}} \|\varphi_E(x) - \varphi_E(y)\|_2 \leq \|\varphi_E(y) - \text{NN}_{k,\varphi_E}(y,\mathcal{R})\|_2\}|}{|\mathcal{G}|}, \tag{9}$$

$$\text{Recall}_{k,\mathcal{R}}(\mathcal{G}) = \frac{|\{y \in \mathcal{R} : \exists_{x\in\mathcal{G}} \|\varphi_E(y) - \varphi_E(x)\|_2 \leq \|\varphi_E(x) - \text{NN}_{k,\varphi_E}(x,\mathcal{G})\|_2\}|}{|\mathcal{R}|}, \tag{10}$$

where $\text{NN}_{k,\varphi_E}(x,\mathcal{G})$ is an embedding of the k-th nearest neighbor of $x$ from the set $\mathcal{G}$ under the embedding function $\varphi_E$.

**Authenticity** measures the fraction of generated samples that appear to be sampled from the approximated distribution, rather than being memorized and recalled training examples, i.e., near-copied (Alaa et al., 2022). Intuitively, generated samples that are closer to a training example than any other training data are considered memorized and therefore not authentic. Formally, for $x \in \mathcal{G}$ we define it's closest neighbor $y^*(x)$ from the reference set $\mathcal{R}$ as

$$y^*(x) = \text{argmin}_{y\in\mathcal{R}} \|\varphi_E(x) - \varphi_E(y)\|_2,$$

and define the Authenticity metric as

$$\text{Authenticity}_\mathcal{R}(\mathcal{G}) = \frac{|\{x \in \mathcal{G} : \|\varphi_E(x) - \varphi_E(y^*(x))\|_2 > \inf_{y\in\mathcal{R}\setminus\{y^*(x)\}}\{\|\varphi_E(y) - \varphi_E(y^*(x))\|_2\}\}|}{|\mathcal{G}|}. \tag{11}$$

**Vendi Score** is a reference-free diversity metric (Friedman and Dieng, 2023). It is defined as the exponential of the Shanon entropy of the trace-normalized similarity matrix's eigenvalues. The similarity matrix is induced by a positive semi-definite kernel function $\kappa$. Formally, for $\mathcal{G} = \{x_1, x_2, \ldots, x_n\}$ and $\mathbf{z} = (z_1, z_2, \ldots, z_n) = (\varphi_E(x_1), \varphi_E(x_2), \ldots, \varphi_E(x_n))$, let $S_{ij}(\mathcal{G}) = \kappa(z_i, z_j)$ be the similarity matrix of set $\mathcal{G}$ with respect to some embedding model E and similarity kernel $\kappa$. Given the eigenvalues of the trace-normalized similarity matrix

$$\overline{\sigma}(S(\mathcal{G})) = \sigma\left(\frac{S(\mathcal{G})}{\text{Tr}(S(\mathcal{G}))}\right) = \{\lambda_1, \lambda_2, \ldots, \lambda_n\},$$

the Vendi Score is defined as

$$\text{VENDI}_\kappa(\mathcal{G}) = \exp\left(-\sum_{i=1}^{n} \lambda_i \log \lambda_i\right), \tag{12}$$

where a higher score indicates more modes in the embedding space and therefore greater diversity. Note that $\sum_{\lambda \in \sigma(A)} \lambda = \text{Tr}(A)$, hence the normalized eigenvalues $\overline{\sigma}(A)$ sum up to 1 and the entropy is well defined.

The usual choice of $\kappa$ is the Gaussian kernel $\kappa(x,y) = \exp(-\|x-y\|_2^2/(2\sigma^2))$ for some $\sigma > 0$. We can approximate the spectrum of the similarity matrix $\sigma(S(\mathcal{G}))$ using Random Fourier Features of kernel $\kappa$. For a Gaussian kernel, fix a set of random feature vectors $\{\omega_1, \omega_2, \ldots, \omega_m\} \sim \mathcal{N}(0, \sigma^{-2} I_{d \times d})$ and define the Random Fourier Feature map $\phi : \mathbb{R}^d \to \mathbb{R}^{2m}$ as

$$\phi(z) = \sqrt{\frac{1}{m}} \left[\cos(\langle \omega_1, z \rangle), \sin(\langle \omega_1, z \rangle), \ldots, \cos(\langle \omega_m, z \rangle), \sin(\langle \omega_m, z \rangle)\right].$$

The resulting similarity matrix $\phi(\mathbf{z})\phi(\mathbf{z})^T = \langle \phi(z_i), \phi(z_j)\rangle_{ij} \in \mathbb{R}^{n \times n}$ has approximately the same spectrum as the original matrix $S_{ij}(\mathcal{G}) = \kappa(z_i, z_j)$ but conveniently shares the nonzero eigenvalues with the matrix $\phi(\mathbf{z})^T \phi(\mathbf{z}) \in \mathbb{R}^{2m \times 2m}$ which is beneficial considering computational savings when $2m < n$. This approximation was introduced in Ospanov et al. (2024). Denote the normalized eigenvalues of approximated similarity matrix $\{\tilde{\lambda}_1, \tilde{\lambda}_2, \ldots, \tilde{\lambda}_{2m}\} = \overline{\sigma}(\phi(\mathbf{z})^T \phi(\mathbf{z}))$ and define the Fourier Kernel Entropy Approximation (FKEA) of the Vendi Score as

$$\text{FKEA-VENDI}(\mathcal{G}) = \exp\left(-\sum_{i=1}^{n} \tilde{\lambda}_i \log \tilde{\lambda}_i\right). \tag{13}$$

Finally, one can generalize the above measure by using the more general Rényi entropy in the exponent

$$\text{FKEA-VENDI}_\alpha(\mathcal{G}) = \exp\left(\frac{1}{1-\alpha} \log\left(\sum_{i=1}^{n} \tilde{\lambda}_i^\alpha\right)\right). \tag{14}$$

Note that $\lim_{\alpha \to 1} \text{FKEA-VENDI}_\alpha(\mathcal{G}) = \text{FKEA-VENDI}(\mathcal{G})$.

### 1.2.3 Metrics from sequence properties

In this group, metrics use representation functions $\varphi_P$ that assign some property to each element $x \in \mathcal{D}$. The metrics are computed using different aggregations of properties given by $\varphi_P$. We denote these metrics as property-based metrics.

**Identity** is defined for scalar properties $\varphi_P : \mathcal{D} \to \mathbb{R}$ and simply aggregates the property with mean and variance

$$\text{Identity}_{\varphi_P}(\mathcal{G}) = \left[\frac{1}{|\mathcal{G}|} \sum_{x \in \mathcal{G}} \varphi_P(x), \frac{1}{|\mathcal{G}|} \sum_{x \in \mathcal{G}} \left(\varphi_P(x) - \frac{1}{|\mathcal{G}|} \sum_{x' \in \mathcal{G}} \varphi_P(x')\right)^2\right]. \tag{15}$$

**Threshold** is defined for scalar properties $\varphi_P : \mathcal{D} \to \mathbb{R}$ and computes the fraction of generated elements whose property exceeds a given threshold

$$\text{Threshold}_{\varphi_P, c}(\mathcal{G}) = \frac{|\{x \in \mathcal{G} : \varphi_P(x) > c\}|}{|\mathcal{G}|}. \tag{16}$$

**Hit-rate** is defined for binary properties $\varphi_P : \mathcal{D} \to \{0, 1\}$ and computes the fraction of positively-labeled elements

$$\text{Hit-rate}_{\varphi_P}(\mathcal{G}) = \frac{|\{x \in \mathcal{G} : \varphi_P(x) = 1\}|}{|\mathcal{G}|}. \tag{17}$$

**Hypervolume** is used in multi-objective optimization, when several and often conflicting properties are optimized (Zitzler and Thiele, 1999). The Hypervolume Indicator (HVI) is defined for positive vector-valued properties $\varphi_P : \mathcal{D} \to \mathbb{R}_+^k$ and measures the volume under the Pareto front of the set $\varphi_P(\mathcal{G})$

$$\text{HVI}_{\varphi_P}(\mathcal{G}) = \text{vol}_{\mathbb{R}_+^k} \left( \text{Pareto}(\varphi_P(\mathcal{G})) \right) = \text{vol}_{\mathbb{R}_+^k} \left( \bigcup_{x \in \mathcal{G}} \{ y \in \mathbb{R}_+^k : \forall_{i \in \{1,2,\ldots,n\}} y_i \leq \varphi_P(x)_i \} \right). \tag{18}$$

Alternatively, one can measure the volume of the convex hull

$$\text{Convex}(\varphi_P(\mathcal{G})) = \left\{ \sum_{g \in \mathcal{G}} \lambda_g g : \lambda_g \geq 0, \sum_{g \in \mathcal{G}} \lambda_g = 1 \right\}.$$

**Conformity score** can be defined for any property function $\varphi_P : \mathcal{D} \to Y$. It uses a conformity measure $A$ which jointly assigns the conformity scores $a$ for each pair $(x, \varphi_P(x))$ given the whole set $\mathcal{G}$. We only require $A$ to be permutation equivariant. Formally, for $\mathcal{G} = \{x_1, x_2, \ldots, x_n\}$ and $\mathcal{R} = \{y_1, y_2, \ldots, y_m\}$, let

$$A\left((x_i, \varphi_P(x_i))_{i=1}^n\right) = (a_i)_{i=1}^n,$$
$$A\left((y_i, \varphi_P(y_i))_{i=1}^m\right) = (b_i)_{i=1}^m,$$

denote the conformity scores of the generated set and reference set, respectively. Since $A$ is permutation equivariant, each $x_i$ has a unique score assigned $a_i = a(x_i)$ which depends only on the value of $x$ and not the index $i$. Similarly $b_i = b(y_i)$.

The Conformity Score metric is defined as the fraction of generated-reference pairs for which the conformity score of the generated sample is greater than the conformity score of the reference sample

$$\text{CS}_{A,\mathcal{R}}(\mathcal{G}) = \frac{|\{(x, y) \in \mathcal{G} \times \mathcal{R} : a(x) \geq b(y)\}|}{|\mathcal{G}||\mathcal{R}|}. \tag{19}$$

In practice, $A$ is the log-likelihood of the joint density over a user-specified set of sequence properties, and the Conformity Score is averaged over $K$-folds sampled from the reference set (Frey et al., 2024).

**KL-divergence** between the property distributions of the reference set and the generated set can be computed for $\varphi_P : \mathcal{D} \to Y$ where $Y$ is either a discrete space or a vector space $\mathbb{R}^k$. Using a kernel density estimator of choice, we approximate the distributions $\lambda_\mathcal{G}$ and $\lambda_\mathcal{R}$

$$\varphi_P(\mathcal{G}) \sim \lambda_\mathcal{G} = p_\mathcal{G} \lambda,$$
$$\varphi_P(\mathcal{R}) \sim \lambda_\mathcal{R} = p_\mathcal{R} \lambda,$$

which are absolutely continuous with respect to some common measure $\lambda$. Then we use the usual definition of Kullback–Leibler divergence

$$\text{KL}_\mathcal{R}(\mathcal{G}) = D_{\text{KL}}(\lambda_\mathcal{G} \| \lambda_\mathcal{R}) = \int_Y p_\mathcal{G} \log \frac{p_\mathcal{G}}{p_\mathcal{R}} d\lambda. \tag{20}$$

In practice, the KL-divergence is approximated using Monte-Carlo sampling.

## 1.3 Diagnostics

Embedding-based metrics require an embedding model to map sequences to fixed-size vector representations. To assess how well an embedding model's sequence representations align with the sequence features of interest, we include two diagnostics for evaluating the embedding models.

### 1.3.1 KNN feature-alignment score

The k-nearest neighbor (KNN) feature-alignment score (Rissom et al., 2025) evaluates an embedding model's alignment with labels assigned by a discrete valued function $y : \mathcal{R} \to \{1, 2, \ldots, K\}$. Let $E : \mathcal{D} \to \mathbb{R}^d$ denote the embedding model mapping a sequence to a fixed-size vector representation and $\text{NN}_{k,E}(x)$ denote the set of $k$ nearest neighbors under embedding E. The KNN Feature Alignment Score of the embedding model E is defined as

$$\text{FAS}_{k,\mathcal{R}}(\text{E}) = \frac{1}{|\mathcal{R}|} \sum_{x \in \mathcal{R}} \frac{|\{x' \in \text{NN}_{k,E}(x) : y(x') = y(x)\}|}{k}. \tag{21}$$

The KNN feature-alignment score yields a value between 0 and 1. An embedding model aligning well with the sequence feature(s) of interest has a value close to 1.

### 1.3.2 Spearman alignment score

The Spearman alignment score (Rissom et al., 2025) evaluates an embedding model's alignment with sequence properties when the properties are continuous. The Spearman alignment score is defined as the Spearman correlation between the pairwise distance between sequences in the embedding space and the feature space, respectively.

## 1.4 Dependencies and licences

`seqme` has the following dependencies: moocore (LGPL-2.1), modlAMP (BSD 3-Clause) (Müller et al., 2017), scikit-learn (BSD 3-Clause) (Pedregosa et al., 2011), UMAP-learn (BSD 3-Clause) (McInnes et al., 2018), Transformers (Apache-2.0) (Wolf et al., 2020), scipy (BSD 3-Clause) (Virtanen et al., 2020), PyTorch2 (custom) (Ansel et al., 2024), pylev (none), matplotlib (custom) (Hunter, 2007), numpy (custom) (Harris et al., 2020), pandas (BSD 3-Clause) (The pandas development team), tqdm (MIT).

All dependencies are compatible with `seqme`'s BSD 3-clause license. To satisfy a BSD 3-clause license, models with incompatible licenses are not directly included in `seqme` but are accessible through third-party repositories and act as optional plugins. Likewise, models with dependency conflicts are not directly included in `seqme` but made compatible with `seqme`'s third-party interface, which supports separate virtual Python environments.

# 2 Supplementary Tables

Table 1: Models currently included in `seqme`. * = models included in third-party repositories and ported to adhere to `seqme`'s interface. For an up-to-date list, check https://seqme.readthedocs.io/en/stable/api/models_index.html.

|  | Model | Description |
|---|---|---|
| **Embedding** | ESM-2 (Lin et al., 2022) | Masked language model with protein and peptide checkpoints. |
|  | RNA-FM (Chen et al., 2022) | RNA language model trained on mRNA and ncRNA sequences. |
|  | GENA-LM (Fishman et al., 2025) | A family of language models for long DNA sequences trained on human DNA sequence. |
|  | Hyformer (Izdebski et al., 2025) | Hybrid language model with small molecule, peptide and antimicrobial peptide checkpoint. |
|  | k-mer frequency | Frequency of pre-specified k-mers of a sequence. |
| **Property** | Physico-chemical properties | 10 physico-chemical properties of peptides/proteins. |
|  | Hyformer (Izdebski et al., 2025) | Hybrid language model predicting sequence perplexity and probability a sequence has antimicrobial properties. |
|  | AMPlify (Li et al., 2022)* | Ensemble model predicting the probability a sequence has antimicrobial properties. |
|  | amPEPpy (Lawrence et al., 2020)* | Random forest model predicting the probability a sequence has antimicrobial properties. |
|  | ESM-2 (Lin et al., 2022) | Masked language model for proteins (and proteins) computing a sequences pseudo-perplexity. |
|  | ESMFold (Lin et al., 2022) | Protein language model predicting proteins atom-wise coordinates, i.e., 3D structure. |
|  | GENA-LM (Fishman et al., 2025) | Family of DNA language models finetuned to predict splice sites and promoter regions. |